\documentclass[10pt,twocolumn,letterpaper]{article}

\usepackage{cvpr}              %

\usepackage[dvipsnames]{xcolor}

\usepackage[utf8]{inputenc} %
\usepackage[T1]{fontenc}    %
\usepackage{url}            %
\usepackage{multirow} %
\usepackage{booktabs}       %
\usepackage{amsfonts}       %
\usepackage{nicefrac}       %
\usepackage{microtype}      %
\usepackage{xcolor}         %
\usepackage{graphicx}
\usepackage{amsmath}
\usepackage{xspace}
\usepackage{booktabs}
\usepackage{pifont}

\makeatletter
\DeclareRobustCommand\onedot{\futurelet\@let@token\@onedot}
\def\@onedot{\ifx\@let@token.\else.\null\fi\xspace}

\def\eg{\emph{e.g}\onedot}

\def\vs{\emph{vs}\onedot}

\newcommand{\cmark}{\ding{51}}
\newcommand{\xmark}{\ding{55}}

\newcommand{\figref}[1]{Fig\onedot~\ref{#1}}
\newcommand{\equref}[1]{Eq\onedot~\eqref{#1}}
\newcommand{\secref}[1]{Sec\onedot~\ref{#1}}
\newcommand{\tabref}[1]{Tab\onedot~\ref{#1}}

\newcommand{\modelname}{SPFormer\xspace}
\newcommand{\modelnamefull}{Superpixel Transformer\xspace}

\makeatletter\renewcommand\paragraph{\@startsection{paragraph}{4}{\z@}
	{.25em \@plus1ex \@minus.2ex}{-.5em}{\normalfont\normalsize\bfseries}}\makeatother
\definecolor{cvprblue}{rgb}{0.21,0.49,0.74}
\usepackage[pagebackref,breaklinks,colorlinks,citecolor=cvprblue]{hyperref}

\def\paperID{16903} %
\def\confName{CVPR}
\def\confYear{2024}

\title{\modelname: Enhancing Vision Transformer with Superpixel Representation}

\author{Jieru Mei$^1$ \quad Liang-Chieh Chen$^2$ \quad Alan Yuille$^1$ \quad Cihang Xie$^3$ \vspace{.3em}\\
$^1$Johns Hopkins University \quad
$^2$Bytedance \quad 
$^3$UC Santa Cruz \vspace{-0.3em}
}

\begin{document}
\maketitle
\begin{abstract}

In this work, we introduce \modelname, a novel Vision Transformer enhanced by superpixel representation. Addressing the limitations of traditional Vision Transformers' fixed-size, non-adaptive patch partitioning, \modelname employs superpixels that adapt to the image's content. This approach divides the image into irregular, semantically coherent regions, effectively capturing intricate details and applicable at both initial and intermediate feature levels.

\modelname, trainable end-to-end, exhibits superior performance across various benchmarks.
Notably, it exhibits significant improvements on the challenging ImageNet benchmark, achieving a 1.4\% increase over DeiT-T and 1.1\% over DeiT-S respectively.
A standout feature of \modelname is its inherent explainability. The superpixel structure offers a window into the model's internal processes, providing valuable insights that enhance the model's interpretability. This level of clarity significantly improves \modelname's robustness, particularly in challenging scenarios such as image rotations and occlusions, demonstrating its adaptability and resilience.

\end{abstract}

\section{Introduction}
\label{sec:intro}

Over the past decade, the vision community has witnessed a remarkable evolution in visual recognition systems, from the resurgence of Convolutional Neural Networks (CNNs) in 2012 \citep{krizhevsky2012imagenet} to the cutting-edge innovation of Vision Transformers (ViTs) in 2020 \citep{dosovitskiy2020image}. This progression has instigated a significant shift in the underlying methodology for feature representation learning, transitioning from pixel-based (for CNNs) to patch-based (for ViTs).

Conventionally, pixel-based representations organize an image as a regular grid, allowing CNNs~\citep{he2016resnet, tan2019efficientnet, mehta2021mobilevit, sandler2018mobilenetv2} to extract local detailed features through sliding window operations.
Despite the inductive bias inherent in CNNs, like translation equivariance, aiding their success in effectively learning visual representations, these networks face a challenge in capturing global-range information, typically necessitating the stacking of multiple convolutional operations and/or additional operations~\citep{li2020nas_nonlocal, chen2018deeplab} to enlarge their receptive fields.

\begin{figure*}[t!]
  \centering
  \includegraphics[width=0.7\linewidth]{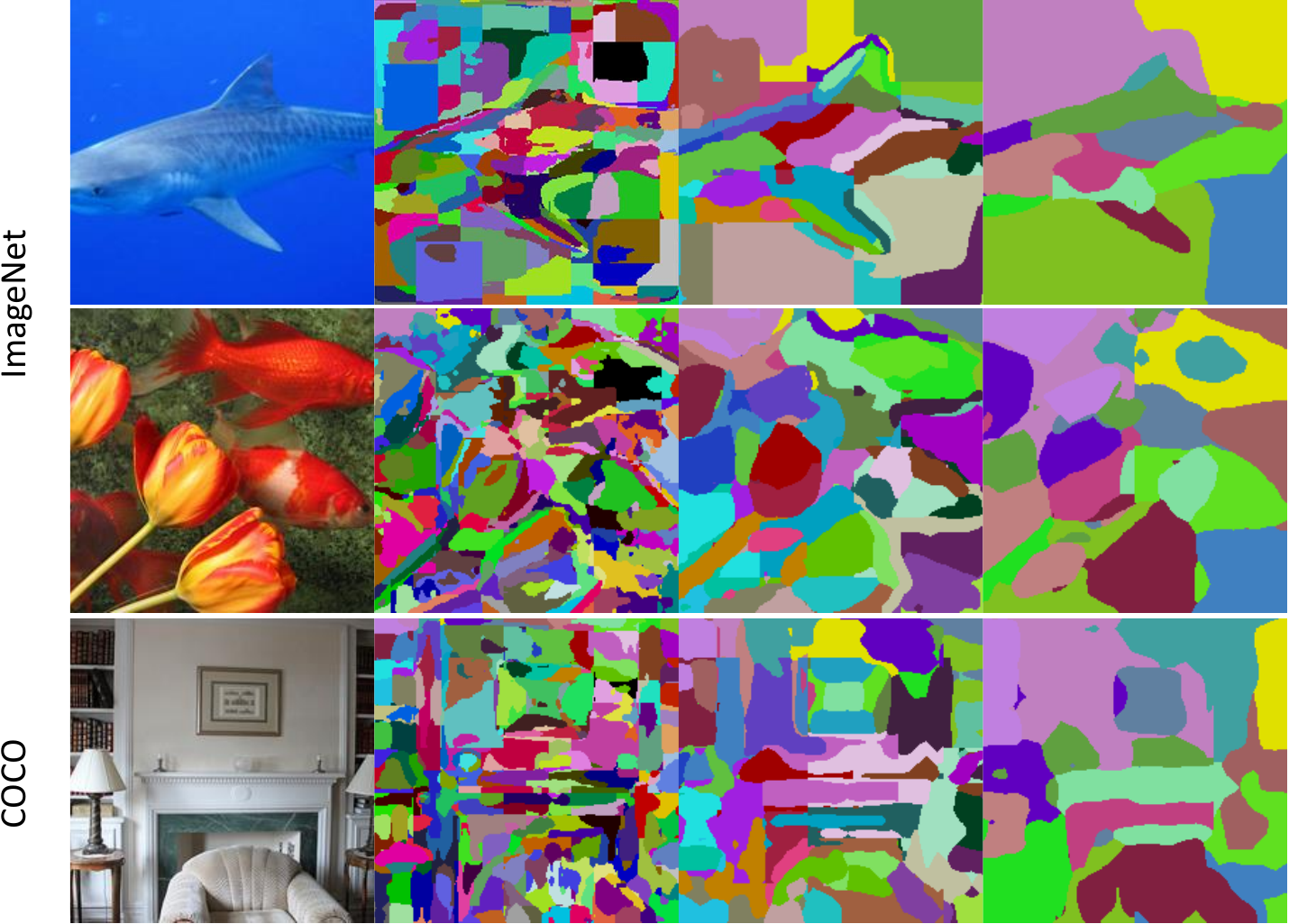}
  \caption{
  Visualization of learned superpixels with our \modelname trained on ImageNet with category labels only.
  For each row, we show input image, visualization of 196, 49, and 16 superpixels. The learned superpixel aligns well with the object boundaries even with 16 superpixels.
  The last row shows results from a COCO image (not trained), demonstrating \modelname's zero-shot ability.
  }
  \label{fig:examples}
\end{figure*}

On the other hand, ViTs~\citep{dosovitskiy2020image} regard an image as a sequence of patches. These patch-based representations, usually of a much lower resolution compared to their pixel-based counterparts, enable global-range self-attention operations in a computationally efficient manner. While the attention mechanism successfully captures global interactions, it does so at the expense of losing local details, like object boundaries. Moreover, the low resolution of patch-based representations poses challenges to adaptation for high-resolution dense prediction tasks such as segmentation and detection, which require both local detail preservation and global context information.

This leads us to ponder an interesting question: \textit{can we derive benefits from both preserved local details and effective long-range relationship capture}? In response, we explore superpixel-based solutions, which have been employed extensively in computer vision prior to the deep learning era~\citep{zhu1996region, shi2000normalized, martin2001database, malik2001contour, borenstein2002class, tu2002image, ren2003learning}. These solutions provide locally coherent structures and reduce computational overhead compared to pixel-wise processing. Specifically, adaptive to the input, superpixels partition an image into irregular regions, with each region grouping pixels with similar semantics. This approach allows for a small number of superpixels, making it amenable to modeling global interactions through self-attention.

Motivated by the inherent limitations of patch representations in ViTs, we introduce an innovative transition to superpixel representation through our Superpixel Cross Attention (SCA). The resulting architecture, \modelnamefull (\modelname), adeptly marries local detail preservation with global-range self-attention, enabling end-to-end trainability. In comparison to standard ViT architectures, \modelname demonstrates remarkable enhancements across various tasks. For instance, it achieves impressive gains on the challenging ImageNet benchmark, such as 1.4\% for DeiT-T and 1.1\% for DeiT-S. Notably, the superpixel representation in \modelname aligns seamlessly with semantic boundaries, even in unseen data. Crucially, the interpretability afforded by our superpixel representation deepens our understanding of the model's decision-making process, elucidating its robustness against rotations and occlusions. These findings highlight the potential of superpixel-based approaches in advancing the field, inspiring future research beyond traditional pixel and patch-based paradigms in visual representation.

\section{Related Work}
\label{sec:related_work}

\paragraph{Pixel Representation}\quad
Convolutional Neural Networks (CNNs)~\citep{lecun1998gradient,krizhevsky2012imagenet,simonyan2015very,szegedy2015going,ioffe2015batch,he2016resnet,tan2019efficientnet,liu2022convnet} process an image as a grid of pixels in a sliding window manner.
CNN has been the dominant network choice since the advent of AlexNet~\citep{krizhevsky2012imagenet}, benefiting from several design choices, such as  translation equivariance and the hierarchical structure to extract multi-scale features.
However, it requires stacking several convolution operations to capture long-range information~\citep{simonyan2015very,he2016resnet}, and it could not easily capture global-range information, as the self-attention operation~\citep{vaswani2017attention}.

\paragraph{Patch Representation}\quad
The self-attention mechanism~\citep{bahdanau2014neural} of Transformer architectures~\citep{vaswani2017attention} effectively captures long-range information.
However, its computation cost is quadratic to the number of input tokens.
Vision Transformers (ViTs)~\citep{dosovitskiy2020image} alleviate the issue by tokenizing (or patchifying) the input image with a sequence of patches (\eg, patch size $16\times16$).
The patch representation~\citep{dosovitskiy2020image} unleashes the power of Transformer architectures~\citep{vaswani2017attention} in computer vision, significantly impacting multiple visual recognition tasks~\citep{russakovsky2015imagenet, carion2020end, zhu2020deformable, wang2021max, hugo2021deit, radford2021learning, bao2022beit, he2022masked, yu2022metaformer, yu2023emergence}.
Due to the lack of the built-in inductive biases as in CNNs, learning with ViTs requires special training enhancements, \eg, large-scale datasets~\citep{sun2017revisiting}, better training recipes~\citep{hugo2021deit, steiner2021train}, or architectural designs~\citep{liu2021swin, wang2021pyramid}.
To mitigate the issue, a few works exploit convolutions~\citep{lecun1998gradient, sandler2018mobilenetv2} to tokenize the images, resulting in hybrid CNN-Transformer architectures~\citep{wu2021cvt, yuan2021incorporating, dai2021coatnet, xiao2021early, mehta2021mobilevit, guo2022cmt, tu2022maxvit, yang2023moat}.
Unlike those works that simply gather knowledge from existing CNNs and ViTs, we explore a different superpixel representation in ViTs.

\paragraph{Superpixel Representation}\quad
Before the deep learning era, superpixel is one of the most popular representations in computer vision~\citep{zhu1996region, shi2000normalized, martin2001database, malik2001contour, borenstein2002class, tu2002image, ren2003learning}.
Ren and Malik~\citep{ren2003learning} preprocess images with superpixels that are locally coherent, preserving the structure necessary for the following recognition tasks.
It also significantly reduces the computation overhead, compared to the pixel-wise processing.
The superpixel clustering methods include graph-based approaches~\citep{shi2000normalized, felzenszwalb2004efficient}, mean-shift~\citep{comaniciu2002mean, vedaldi2008quick}, or k-means clustering~\citep{lloyd1982least, achanta2012slic}.
Thanks to its effective representation, recently some works attempt to incorporate clustering methods into deep learning frameworks~\citep{Jampani2018ssn, yang2020superpixel, locatello2020object, xu2022groupvit, yu2022cmt, zhang2022semantic, yu2022kmeans_transformer, ma2023image, huang2023stvit, zhu2023superpixel}.
For example, SSN~\citep{Jampani2018ssn} integrates the differentiable SLIC~\citep{achanta2012slic} to CNNs, allowing end-to-end training.
\citet{yu2022cmt, yu2022kmeans_transformer} regard object queries~\citep{carion2020end, wang2021max} as cluster centers in Transformer decoders~\citep{vaswani2017attention}.
SViT~\citep{huang2023stvit} clusters the tokens to form the super tokens, where the clustering process has no gradient passed through\footnote{From official code: \url{https://github.com/hhb072/STViT/blob/main/models/stvit.py\#L206}}.
Consequently, their network is not aware of the clustering process and could not recover from the clustering error.
CoCs~\citep{ma2023image} groups pixels into clusters, while aggregating features within each cluster by regarding the image as a set of points with coordinates concatenated.
In contrast, our proposed method groups pixels into superpixels, and models their global relationship via self-attention. 
Furthermore, during clustering, CoCs uses a Swin-style window partition~\citep{liu2021swin} that introduces visual artifacts, especially around the window boundaries.

\section{Method}
\label{sec:method}

In this section, we formalize our superpixel representation and compare it with traditional methods in \secref{sec:representation}. We then detail the integration of this representation with our Superpixel Cross Attention mechanism in \secref{sec:block}. Building on these concepts, our model \modelname, which exemplifies an explainable and efficient approach to image processing, is presented in \secref{sec:arch}.

\subsection{Superpixel Representation: Bridging Pixel and Patch Approaches}
\label{sec:representation}

In the evolving landscape of feature representation, the transition from pixel to patch-based methods in Vision Transformers has opened new avenues for image processing. However, each method has limitations, inspiring our exploration of a more adaptive and efficient representation: superpixels.

\paragraph{Pixel Representation}
Conventional pixel representation treats an image $\mathbf{I}$ as a grid of high-resolution pixels, with $\mathbf{I} \in \mathcal{R}^{c \times h \times w}$. This representation, dominant in CNN-based methods, suffers from restricted contextual integration due to limited receptive fields. While self-attention mechanisms could theoretically enhance this integration, their application at this resolution is computationally burdensome due to the quadratic complexity with respect to the number of pixels.

\paragraph{Patch Representation}
Vision Transformers typically use a lower resolution patch representation, $\mathbf{P} \in \mathcal{R}^{c \times p_h \times p_w}$, reducing input length significantly. This reduction facilitates the application of self-attention mechanisms but at the cost of finer details and contextual nuances due to the coarse granularity of patches.

\paragraph{Superpixel Representation}
Our superpixel representation synthesizes the detail of pixel-based methods with the efficiency of patch-based approaches, consisting of superpixel features $\mathbf{S} \in \mathcal{R}^{c \times s_h \times s_w}$ and pixel-to-superpixel associations $\mathbf{A} \in \mathcal{R}^{n \times h \times w}$.

1. \textbf{Neighboring Superpixels ($\mathcal{N}_i$):} For each pixel $i$, neighboring superpixels are defined as $\mathcal{N}_i$, including the nearest superpixel and its Moore Neighborhood for $n=9$. The association matrix $\mathbf{A}$ delineates relationships between pixels and their neighboring superpixels, fostering a competitive dynamic among them.

2. \textbf{Superpixel's Local Window ($\mathcal{W}_p$):} Derived from the neighboring superpixels ($\mathcal{N}_i$), each superpixel $p$ is associated with a local window $\mathcal{W}_p$. This window encompasses the neighboring pixels that contribute to the superpixel's feature representation. The overlapping nature of these local windows is pivotal for the implementation of SCA with a sliding window approach in \secref{sec:block}, enabling more nuanced attention mechanisms.

The transformation from superpixel to pixel representation is captured by:
\begin{equation}
\mathbf{I}_i = \sum_{p \in \mathcal{N}_i} \mathbf{A}_{ip} \cdot \mathbf{S}_p
\label{eq:pixelify}
\end{equation}
where $\mathbf{I}_i$ signifies the feature of the $i$-th pixel. This approach ensures boundary information preservation and finer granularity compared to direct patch upsampling, translating from a dense pixel grid to a coarser superpixel grid.

The superpixel representation uniquely conserves boundary information, enabling the maintenance of high-resolution features crucial for detailed tasks. Its robustness to image distortions, such as rotation and occlusion, makes it robust compared to traditional pixel or patch methods. In summary, superpixel representation is efficient due to its reduced resolution, explainable through semantic pixel clustering, and robust against challenging image transformations.

\subsection{Superpixel Cross Attention}
\label{sec:block}

\begin{figure}[!t]
  \centering
  \includegraphics[width=\linewidth]{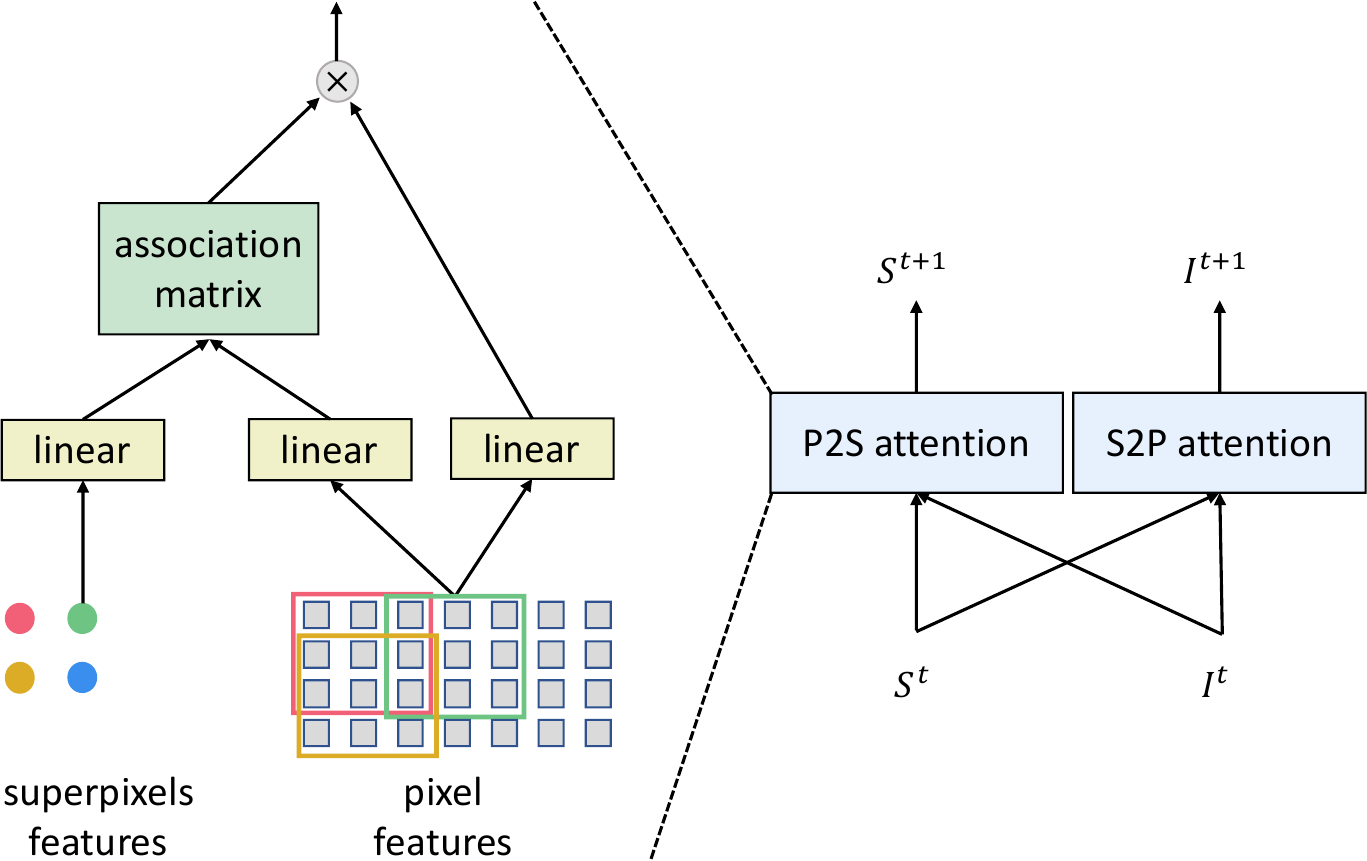}
  \caption{
  Illustration of our SCA module for iterative refinement of both superpixel and pixel features using a sliding window-based cross-attention mechanism. Each superpixel cross-attends to a localized region of pixels, as highlighted in the colored rectangle. On the left, we detail the Pixel-to-Superpixel (P2S) cross-attention process, while the Superpixel-to-Pixel (S2P) cross-attention is depicted similarly, albeit with reversed roles for superpixel and pixel.
  }
  \label{fig:superpixel_block}
\end{figure}

Given an initial pixel representation $\mathbf{I}^0$ and superpixel features $\mathbf{S}^0$, our method iteratively updates these features. At each iteration $t$, both superpixel features $\mathbf{S}^t$ and the association $\mathbf{A}^t$ are refined using a cross-attention mechanism within a sliding window, as depicted in \figref{fig:superpixel_block}. This mechanism is designed to maintain the locality of superpixels while ensuring high computational efficiency.

The SCA module, pivotal to our approach, encompasses two types of cross-attention: Pixel-to-Superpixel (P2S) and Superpixel-to-Pixel (S2P).
For the P2S cross-attention, superpixel features cross-attend to pixel features within a localized region, enhancing the superpixel representations. Conversely, in the S2P cross-attention, pixel features are updated based on neighboring superpixels, refining the pixel representation.

The P2S cross-attention updates the superpixel features $\mathbf{S}^t$ by aggregating relevant pixel features, computed as:
\begin{equation}
\mathbf{S}^t_p = \mathbf{S}^{t-1}_p + \sum_{i \in \mathcal{W}_p} \text{softmax}_i \left( \mathbf{q}_{\mathbf{S}^{t-1}_p} \cdot \mathbf{k}_{\mathbf{I}^{t-1}_i} \right) \mathbf{v}_{\mathbf{I}^{t-1}_i}
\label{eq:superpixel_update}
\end{equation}
Here, $\mathcal{W}_p$ indicates the set of pixels within the local window of superpixel $p$. The vectors $\mathbf{q}$ (query), $\mathbf{k}$ (key), and $\mathbf{v}$ (value) are derived from linear transformations of the prior iteration's superpixel features $\mathbf{S}^{t-1}_p$ and pixel features $\mathbf{I}^{t-1}_i$.

In the S2P cross-attention, pixel features $\mathbf{I}^t$ are updated using the updated associations $\mathbf{A}^t_{ip}$, calculated as follows:
\begin{equation}
\mathbf{A}^t_{ip} = \text{softmax}_{p \in \mathcal{N}_i} \left( \mathbf{q}_{\mathbf{I}^{t-1}_i} \cdot \mathbf{k}_{\mathbf{S}^{t-1}_p} \right)
\end{equation}
where $\mathcal{N}_i$ denotes the neighboring superpixels of pixel $i$. The updated pixel representation $\mathbf{I}^t_i$ is then derived by:
\begin{equation}
\mathbf{I}^t_i = \mathbf{I}^{t-1}_i + \sum_{p \in \mathcal{N}_i} \mathbf{A}^t_{ip} \cdot \mathbf{v}_{\mathbf{S}^{t-1}_p}
\label{eq:pixel_update}
\end{equation}
Here, the value vector $\mathbf{v}$ is obtained through a linear transformation of the preceding superpixel features $\mathbf{S}^{t - 1}_p$.

To incorporate positional information within SCA, we utilize Convolution Position Embedding (CPE)~\citep{huang2023stvit}, which captures the spatial relationships within the image. Prior to applying P2S and S2P cross-attentions, both superpixel and pixel features are augmented with CPE, implemented as a $3\times3$ depthwise convolution with a skip connection. This enhancement strengthens the association between pixels and superpixels based on their spatial proximity, fostering more accurate pixel-superpixel alignments.

These iterative update equations are fundamental to refining both pixel and superpixel feature representations, thereby enhancing the overall quality and accuracy of the feature representation. The proposed SCA module, capable of multiple iterations $t$, is a cornerstone of our method and will be further elaborated in the subsequent architecture \modelname in the next subsection.

\subsection{\modelname Architecture}
\label{sec:arch}

Our architecture, designed to leverage the advantages of the proposed superpixel representation, introduces minimal alterations from the standard ViT~\citep{dosovitskiy2020image}. In alignment with the ViT methodology, we employ a non-overlapping patchify layer. However, we utilize a smaller window size of $4\times4$ for extracting initial pixel features, as opposed to the conventional $16\times16$.  This reduction in window size is made feasible by the effective superpixel representation, which significantly decreases the input length.

\begin{figure*}[h]
  \centering
  \includegraphics[width=0.75\linewidth]{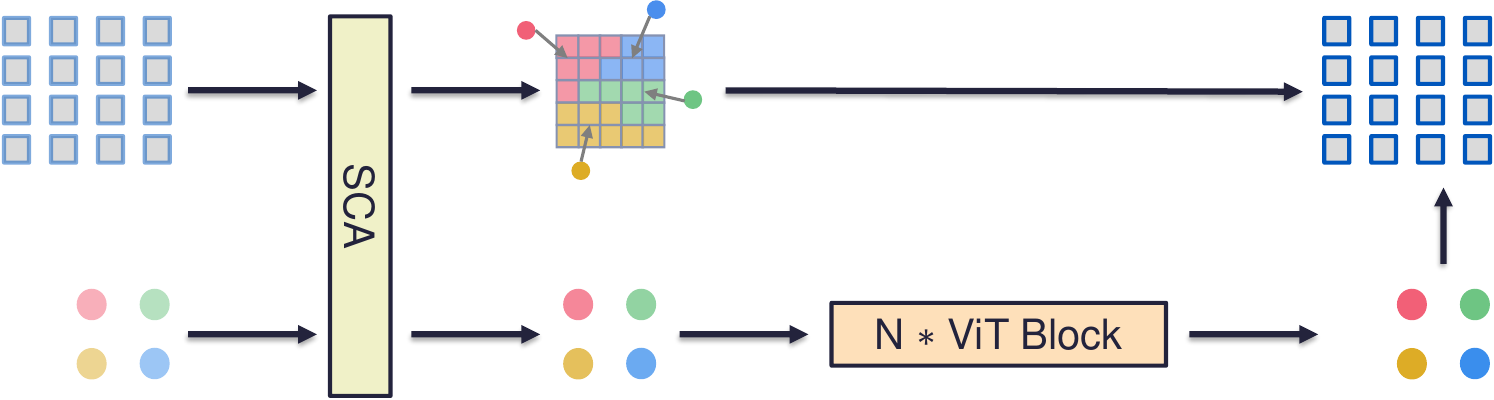}
  \caption{
  Illustration of a single stage of the \modelname architecture. It starts with initial superpixel features and pixel features as inputs. The SCA module iteratively refines superpixel features, enhancing their semantic richness. These features are then processed by the Multi-Head Self-Attention (MHSA) for global contextual understanding. The stage concludes by updating the pixel features based on the enriched superpixel information, readying them for the next stage or for final pooling and classification. This design showcases the efficient integration of local detail and global context in \modelname.
  }
  \label{fig:arch}
\end{figure*}

Initially, the superpixel features $\mathbf{S}^0$ are derived using a $1 \times 1$ convolution and $4\times4$ average pooling, based on the pixel features $\mathbf{I}^0$. We utilize the SCA module to iteratively update these superpixel features $\mathbf{S}^0$ and the association $\mathbf{A}^0$ (across $t$ iterations), as outlined in \secref{sec:block}. The SCA module capitalizes on the local spatial context within a superpixel to enhance its representation. Subsequently, the updated superpixel features undergo Multi-Head Self-Attention (MHSA), enabling the network to discern long-range dependencies and contextual information across various superpixels, thus facilitating a comprehensive understanding of the image.

We observed that, even with multiple iterations (e.g., $t > 2$) within an SCA module, generating lower-level superpixel representations may not perfectly align with the overall context, mainly due to insufficient semantic information. To address this, we propose a gradual refinement strategy for the superpixel representations through multiple SCA modules, each comprising a few iterations (e.g., $t=2$). Each subsequent SCA module utilizes the updated pixel features to generate semantically richer superpixels.

Specifically, prior to advancing to the next SCA module, the superpixel features are projected through a $1 \times 1$ convolution. The pixel features are then updated according to \equref{eq:pixelify}, incorporating a skip connection~\citep{he2016resnet}. This method ensures that pixel representations are refined in light of the globally context-enhanced superpixel features obtained from preceding SCA and MHSA modules. Rather than reinitializing superpixel features from the beginning, we utilize the contextually enriched superpixel features from the preceding stage as the starting point. This methodology, depicted in \figref{fig:arch}, systematically enhances the superpixel representations, allowing for the capture of progressively more complex semantic information.

In the concluding stage, global average pooling is applied to the superpixel features, and the resultant representation is fed into a linear classifier for image classification tasks.

Conceptually, our network architecture can be envisaged as a dual-branch structure. One branch maintains a dense pixel representation with high resolution, while the other branch is dedicated to our proposed low-resolution superpixel representation. Minimal direct operations are performed on the dense pixel representation, allowing us to concentrate most computational efforts on the more efficient superpixel representation. This dual-branch approach achieves computational efficiency without compromising the preservation of local details in the image representation.

\section{Experiments}
\label{sec:exp}

We first detail our method's implementation in \secref{sec:implement_details}. We then demonstrate its explainability in \secref{sec:explainability} and assess efficiency on image classification and segmentation in \secref{sec:efficiency}.

\subsection{Implementation Details}
\label{sec:implement_details}

\modelname establishes a specific ratio between the dimensions of superpixel and pixel features. By design, the spatial dimensions of superpixel features are reduced to $\frac{1}{4}\times\frac{1}{4}$ of their corresponding pixel features. This downscaling strategy is instrumental in encoding contextually rich information at a more abstract level, while simultaneously preserving essential details at the pixel level.

\begin{figure}[htbp]
  \centering
  \includegraphics[width=1.0\linewidth]{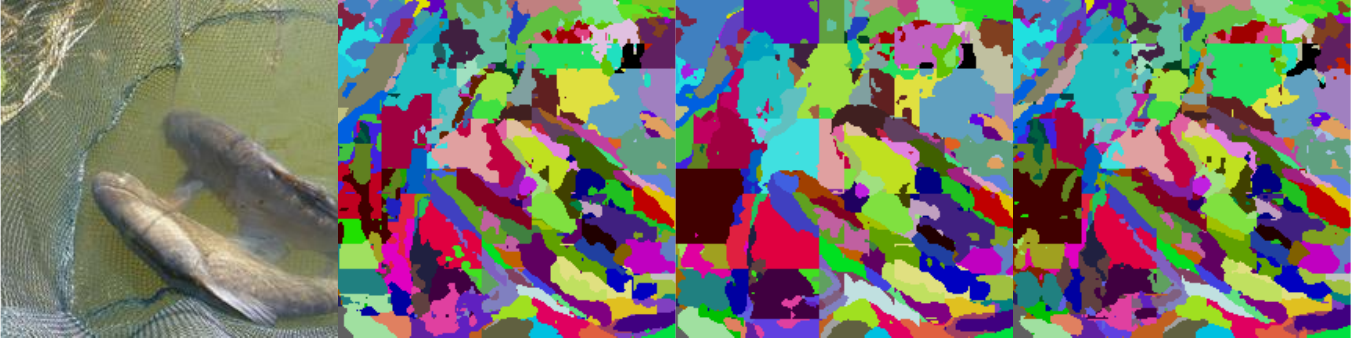}
  \caption{The multi-head SCA design generates multiple superpixel representations, each capturing different semantic relationships and addressing the ambiguity in superpixel over-segmentation.}
  \label{fig:multihead}
\end{figure}

In the SCA module, we employ multi-head attention to manage attention and interaction between superpixels effectively. This setup not only leverages global contextual information optimally but also produces multiple superpixel representations, as shown in \figref{fig:multihead}. These varied representations capture different granularities, addressing the ambiguity commonly associated with superpixel over-segmentation. Specifically, we allocate two heads for our smaller variants (\modelname-T and \modelname-S) and three heads for the base model (\modelname-B).

The SCA blocks are integrated into the standard ViT architecture, strategically positioned just before the first and third self-attention blocks. We adopt the LayerScale technique, as described in \citet{hugo2021cait}, to regulate gradient flow, thereby enhancing training stability and convergence. It is noteworthy that, combined with the residual connection as formulated in \equref{eq:superpixel_update} and \equref{eq:pixel_update}, our method initially resembles vanilla patches, evolving to leverage superpixels as training progresses.

Following the training protocols detailed in DeiT \citep{hugo2021deit}, we utilize strong data augmentations, the AdamW optimizer, and a cosine decay learning rate schedule. All models undergo training on the ImageNet dataset \citep{russakovsky2015imagenet} for a duration of 300 epochs. During the training phase of \modelname-B/16, we faced significant overfitting challenges. To counteract this, we increased the Stochastic Depth~\citep{huang2016stochastic_depth} rate from 0.1 to 0.6, effectively mitigating the overfitting. This adjustment underscores the potential need for more sophisticated regularization techniques, especially those tailored to the superpixel representation, which we aim to investigate in future research endeavors.

For \modelname variants, the default configuration employs a $4\times4$ patchify layer. Variants augmented with two convolution layers of kernel size 3 with stride 2 are denoted by $^\dag$. Other configurations, adapting different superpixel sizes, are indicated by their ViT-equivalent patch sizes, such as \modelname/32 for a $32\times32$ patch size. This notation ensures clear distinction between each variant in our experiments.

\subsection{Unveiling \modelname's Explainability}
\label{sec:explainability}

Integrating superpixel representation into the Vision Transformer architecture adds a significant layer of explainability compared to conventional fixed patch partition methods. This section first discusses the inherent explainability of our superpixel representation, followed by an evaluation of its semantic alignment and generalizability to unseen data.

\subsubsection{Superpixel Representation as an Explainability Tool}
\label{sec:superpixel_explainability}

Our method's superpixel representation can be visualized through the association matrix $\mathbf{A}$, providing insights into the model's internal processing. In \figref{fig:examples}, we visualize the learned soft associations by selecting the argmax over the superpixels:

\begin{equation}
\hat{\mathbf{A}} = \operatorname{argmax}(\mathbf{A})
\end{equation}

These visualizations reveal that, even with a soft association, the superpixels generally align with image boundaries. This alignment is noteworthy as it emerges even though the network is only trained on image category labels. Thus, the model segments images into irregular, semantics-aware regions while reducing the number of tokens needed for representation.

\begin{figure}[htbp]
  \centering
  \includegraphics[width=\linewidth]{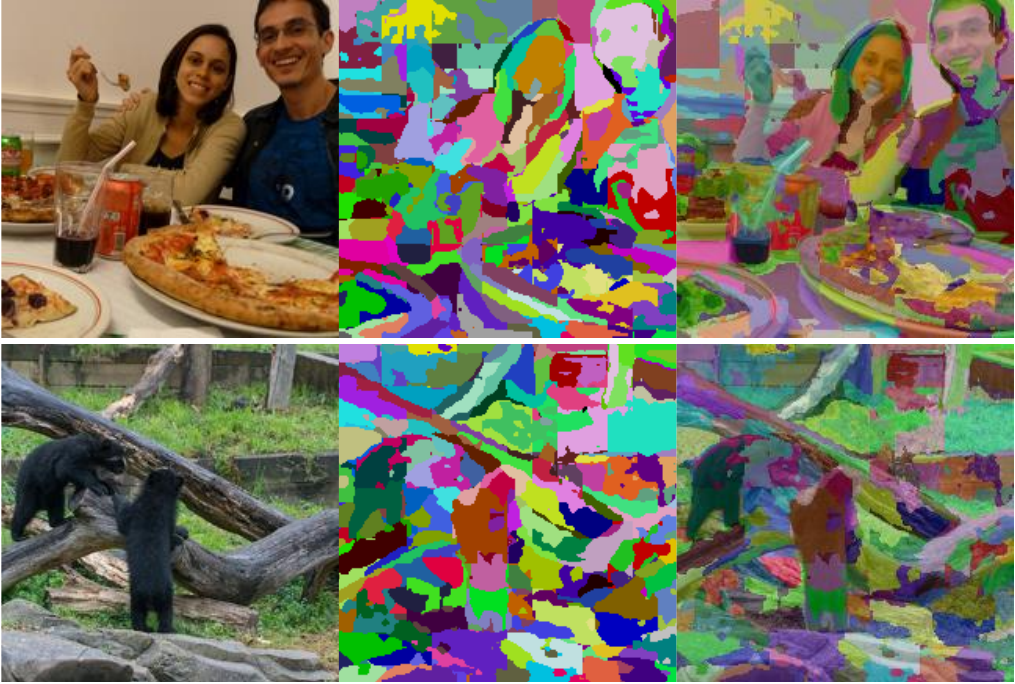}
  \caption{Zero-shot transferability on the COCO dataset. Trained solely on ImageNet, \modelname demonstrates effective segmentation of unseen COCO images into detailed superpixels. 196 superpixels are used in this visualization.}
  \label{fig:ood}
\end{figure}

Furthermore, we assess the generalizability of our superpixel representation using the COCO dataset \citep{lin2014coco}, which consists of high-resolution images with complex scenes. For this evaluation, we resize and center-crop COCO images to align with the ImageNet evaluation pipeline. \figref{fig:ood} showcases the visual representation of superpixels on these images. Remarkably, the superpixels generated by \modelname, trained exclusively on ImageNet, adapt well to this unseen data, capturing intricate structures such as thin objects. This adaptability highlights the model's capability to preserve detail and generalize its superpixel representation to new contexts.

\subsubsection{Semantic Alignment of Superpixels}
\label{sec:semantic_alignment}

Our evaluation of superpixel representation focuses on its ability to align with ground truth boundaries in images, despite the model not being trained on the datasets used for this assessment. This test involved a quantitative analysis on both object and part levels using the Pascal VOC 2012 dataset \citep{everingham2015pascal} and Pascal-Part-58 \citep{zhao2019part58}. Remarkably, these assessments were performed without any training on these specific datasets, underscoring the model's generalization capabilities.

In our approach, each superpixel or patch's prediction is derived by aggregating the ground truth labels of the pixels it encompasses. We assign the most frequently occurring label within a superpixel as its prediction, assuming optimal classification. This method leverages the soft associations produced by our SCA module, where predictions are formed by combining pixel labels with their corresponding weights and upscaled as per \equref{eq:pixelify}.

Diverging from the single-superpixel outcome of traditional patch representation, our model employs a multi-head design in the SCA module. This allows for the creation of multiple, distinct superpixels for each head, enhancing the richness and diversity of the extracted features (see \figref{fig:multihead}). For our evaluations, we computed an average of predictions across all heads. It's noteworthy that effective feature extraction in our model is deemed successful if even a single head accurately identifies a superpixel.

The proficiency of our superpixel approach is demonstrated in its performance compared to vanilla ViTs that utilize patch representations with a stride of 16. ViTs often suffer from a granularity trade-off, losing finer details in favor of broader patch representations. In contrast, the superpixels from our SCA module, as shown in \tabref{tab:superpixel_quality}, manifest substantial improvements — achieving a 4.2\% increase in object-level and 4.6\% in part-level mean Intersection over Union (mIoU) with \modelname-S$^\dag$. Furthermore, these superpixels display a quality comparable to those from traditional superpixel methods like SLIC \citep{achanta2012slic}, highlighting our method's effectiveness in capturing detailed semantic information without direct training on the evaluation datasets.

\begin{table}[htb]
\caption{Evaluation of superpixel quality in a zero-shot setting on Pascal VOC 2012 and Pascal-Parts-58 datasets, using 196 patches/superpixels. Our \modelname variants demonstrate notable improvements over traditional patch representations and are competitive with the SLIC method.}
\centering
\begin{tabular}{lcc|cc}
\toprule
\multicolumn{1}{l}{\multirow{2}{*}{Method}} & \multicolumn{2}{c}{Pascal Voc2012} & \multicolumn{2}{c}{Pascal-Parts-58} \\
\cmidrule{2-3} \cmidrule{4-5}
& mIoU & mAcc & mIoU & mAcc \\
\midrule
Patch & 87.8 & 92.8 & 68.7 & 78.2 \\
\midrule
\modelname-T$^\dag$ & 91.5 & 95.7 & 71.5 & 79.9 \\
\modelname-S$^\dag$ & 92.0 & 96.6 & 73.3 & 82.4 \\
\modelname-B$^\dag$ & 91.2 & 96.3 & 72.5 & 81.4 \\
\midrule
SLIC~\citep{achanta2012slic} & 92.5 & 95.4 & 74.0 & 81.7 \\
\bottomrule
\end{tabular}
\label{tab:superpixel_quality}
\end{table}

\subsubsection{Explainability-Driven Robustness}
\label{sec:robustness}

The robustness of \modelname is deeply intertwined with its explainability, particularly through the superpixel representation. This section explores how the model's transparent and interpretable features contribute to its resilience against image modifications like rotation and occlusion.

\begin{figure}[htb]
  \centering
  \includegraphics[width=\linewidth]{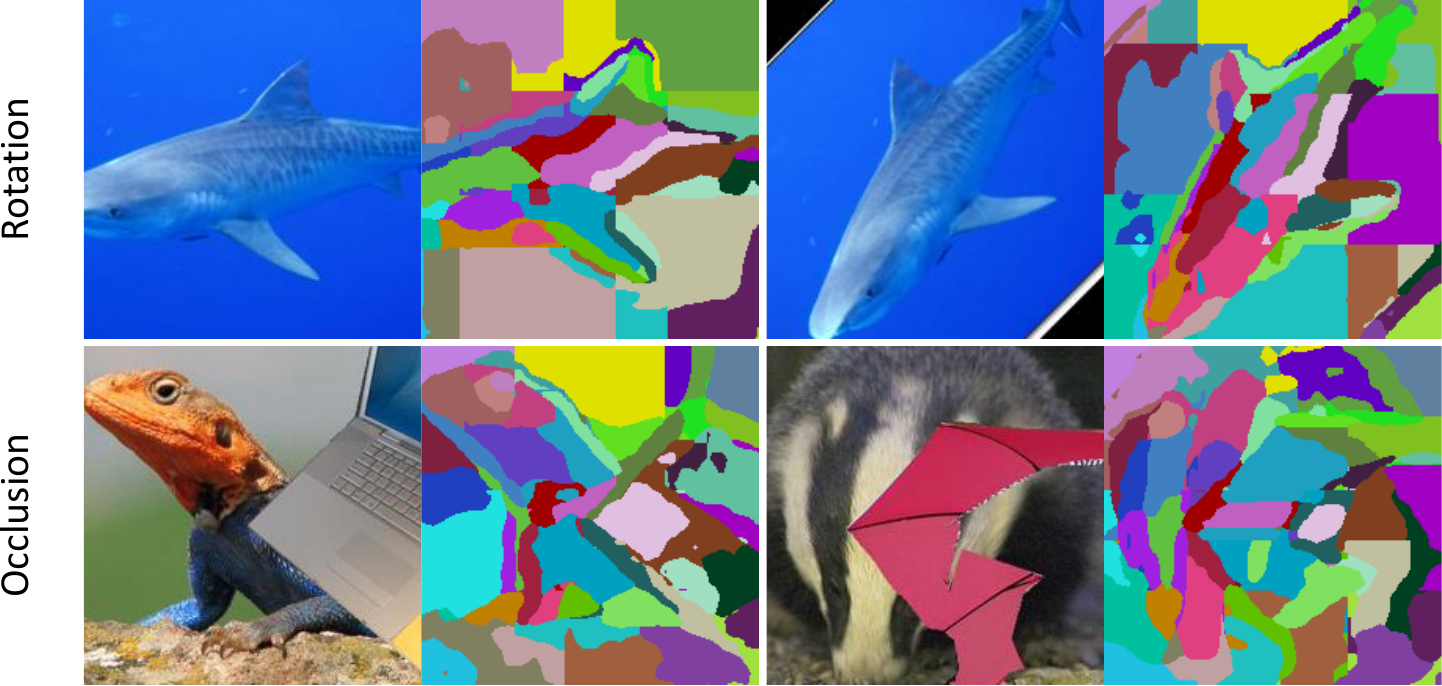}
  \caption{Visualization of \modelname's superpixel representation under rotation and occlusion, highlighting the model's adaptability and robustness.}
  \label{fig:robustness}
  \vspace{-1em}
\end{figure}

\paragraph{Robustness to Rotation}
Our model's capacity to generate coherent superpixels even under rotational transformations showcases the robustness afforded by its explainable structure. By visualizing how superpixels adapt to rotated images, we gain insights into the model's stability in varied orientations. While \modelname demonstrates a heightened robustness to rotation, it still exhibits some limitations, likely due to the learnable absolute position embeddings not being inherently rotation-invariant. These observations suggest potential avenues for enhancing rotational robustness, possibly through integrating rotation-invariant mechanisms within the superpixel representation or the network architecture. Figure~\ref{fig:robustness} illustrates the model's performance under rotation, and Table~\ref{tab:rotation} quantifies this robustness, under patch size 32.

\paragraph{Robustness to Occlusion}
The occlusion robustness of \modelname is another facet where explainability plays a key role. By examining superpixel behavior in occluded images, we observe the model's ability to distinguish between occluders and the object of interest. Unlike traditional patch-based representations, which tend to blend occluders with the object, our superpixel representation more effectively isolates and identifies obscured parts of the image. This nuanced differentiation is a direct result of the model's explainable superpixel structure, which provides a more detailed and context-aware interpretation of the image content, as demonstrated in Figure~\ref{fig:robustness}.

\begin{table}[htb]
\caption{Quantitative evaluation of \modelname's robustness to rotation, comparing performance at different angles.
Variants augmented with two convolution layers of kernel size 3 with stride 2 are denoted by $^\dag$.
}
\centering
\begin{tabular}{l|c|c|c|c}
\toprule
Model & Clean & 15 & 30 & 45 \\
\midrule
DeiT-S/32~\citep{hugo2021deit} & 73.3 & 71.1 & 67.7 & 59.5 \\
\midrule
\modelname-S/32$^\dag$ & 77.9 & 75.2 & 73.4 & 66.9 \\
\bottomrule
\end{tabular}
\label{tab:rotation}
\end{table}

\subsection{Efficiency in Image Classification and Segmentation}
\label{sec:efficiency}

\begin{table}[t!]
\caption{Comparative analysis of \modelname's performance on ImageNet classification against DeiT baselines.
Variants augmented with two convolution layers of kernel size 3 with stride 2 are denoted by $^\dag$.
}
\centering
\begin{tabular}{l|c|c|c}
\toprule
Model & \#Params & \#FLOPs & Top-1 \\
\midrule
\modelname-S/56 & 22M & 0.5G & 72.3 \\
\midrule
DeiT-T~\citep{hugo2021deit} & 5M & 1.3G & 72.2 \\
\modelname-T & 5M & 1.3G & 73.6 \\
\modelname-T$^\dag$ & 5M & 1.3G & 75.0 \\
\midrule
DeiT-S/32~\citep{hugo2021deit} & 22M & 1.1G & 73.3\\
\modelname-S/32 & 22M & 1.2G & 76.4 \\
\modelname-S/32$^\dag$ & 22M & 1.3G & 77.9 \\
\midrule
DeiT-S~\citep{hugo2021deit} & 22M & 4.6G & 79.9 \\
\modelname-S & 22M & 5.2G & 81.0 \\
\modelname-S$^\dag$ & 22M & 5.3G & 81.7 \\
\midrule
DeiT-B~\citep{hugo2021deit} & 87M & 17.5G & 81.8 \\
\modelname-B & 87M & 19.2G & 82.4 \\
\modelname-B$^\dag$ & 87M & 19.2G & 82.7 \\
\bottomrule
\end{tabular}
\label{tab:exp_imagenet}
\vspace{-1em}
\end{table}

\subsubsection{Main Results on ImageNet}
\label{sec:imagenet_results}

Our assessment of \modelname on ImageNet underlines its enhanced efficiency and performance compared to the DeiT baseline across diverse configurations, as shown in \tabref{tab:exp_imagenet}. Notably, \modelname-S, utilizing the standard ViT configuration with 196 tokens, surpasses DeiT-S by a margin of 1.1\% (achieving 81.0\% \vs DeiT's 79.9\%). In the case of \modelname-T, it exceeds DeiT-T by 1.4\% (73.6\% \vs 72.2\%). This advantage becomes more pronounced when larger patch sizes, such as 32, are used. While DeiT-S/32 exhibits a decline in performance due to its coarse granularity, \modelname-S/32 maintains robust performance at 76.4\%, even outperforming DeiT-T by a significant 4.2\% with less FLOPs.

A notable aspect of \modelname is the shift in computational load, with MLPs taking precedence over self-attention mechanisms. This redistribution suggests an alternative scaling strategy, namely increasing image resolution to encapsulate finer details. By adapting to a higher resolution of 448, through doubling the window size from 4 to 8, \modelname retains computational efficiency and achieves a 0.3\% improvement in performance compared to its standard configuration. In contrast, when DeiT-S employs a similar strategy, it gains a marginal improvement of only 0.1\%, limited by the granularity of its patch representation.

Further enhancing \modelname's initial feature extraction phase in the superpixel cross-attention stage, we introduce a lightweight convolution stem comprising two or three $3 \times 3$ convolutions with a stride of 2. This enhancement has consistently improved performance, exemplified by \modelname-S/32$^\dag$, which witnesses an additional increase of 1.5\% in ImageNet accuracy, reaching 77.9\%.

\begin{table}[htb]
\caption{Ablation study on the design choices in \modelname.}
\centering
\resizebox{\linewidth}{!}{
\begin{tabular}{l|c|c|c}
\toprule
Model & \#Params & \#FLOPs & Top-1 \\
\midrule
\modelname-S/32 & 22M & 1.2G & 76.4 \\
\midrule
Single Iteration in SCA & 22M & 1.2G & 75.4 \\
SCA at Initial Layer Only & 22M & 1.2G & 74.8 \\
Single-Head SCA & 22M & 1.2G & 75.6 \\
Learnable Position Embeddings & 22M & 1.2G & 76.1 \\
\bottomrule
\end{tabular}
}
\label{tab:ablations}
\vspace{-1em}
\end{table}

\subsubsection{Ablation Study: Design Choices in \modelname}
\label{sec:imagenet_ablations}

We evaluates key design elements of \modelname-S/32 on the ImageNet validation set. We investigate the impacts of iteration count in the SCA module, the placement of SCA within the architecture, the use of multi-head attention, and the choice of position embeddings.

The findings highlight the importance of multiple iterations in SCA for performance enhancement, with a single iteration leading to a 1.0\% drop in accuracy. The strategic placement of SCA across different layers is crucial, as restricting it to the initial layer causes a 1.6\% accuracy reduction, indicating that higher-level features play a vital role in augmenting semantic depth and in rectifying early-stage superpixel inaccuracies. Furthermore, employing multi-head attention in SCA is significant for capturing diverse superpixel relationships, with its absence leading to a 0.8\% decrease in accuracy. Lastly, using learnable position embeddings over CPE results in a slight drop in performance.

This ablation study validates the effectiveness of the considered design choices in \modelname, affirming their contributions to the overall performance of the model.

\subsubsection{Semantic Segmentation: Utilizing \modelname's High-Resolution Feature Preservation}
\label{sec:semantic_segmentation}

\modelname's superpixel representation intrinsically maintains higher resolution features, making it particularly suitable for semantic segmentation tasks. This characteristic allows for detailed and context-rich segmentation outputs, a key advantage over traditional patch-based methods.

Incorporating \modelname into the SETR \citep{zheng2021setr} framework, we enhance segmentation performance by directly classifying individual superpixels. This direct approach leverages \modelname's high-resolution feature preservation, allowing for more nuanced segmentation. The final segmentation maps are generated by upscaling the superpixel-based logits using \equref{eq:pixelify}.

We evaluate \modelname on the ADE20K \citep{zhou2017ade20k} and Pascal Context \citep{mottaghi2014pascal_context} datasets. Utilizing ImageNet-pretrained models, \modelname demonstrates significant improvements in mIoU, highlighting its effectiveness in detailed segmentation tasks. Detailed training parameters and methodologies for these evaluations are provided in the supplementary material.

As shown in \tabref{tab:semantic_seg} and \tabref{tab:semantic_seg_pascal}, the performance gains in mIoU are noteworthy when using ImageNet-pretrained models: 4.2\% improvement on ADE20K and 2.8\% on Pascal Context. These results not only highlight the detailed nature of \modelname's superpixel representation but also its adaptability to diverse and complex datasets. To further validate the intrinsic segmentation capabilities of \modelname, we conduct additional training from scratch. This approach reiterates the model's strength in maintaining high-resolution features independently of pretraining influences, leading to mIoU improvements of 3.0\% on ADE20K and 3.1\% on Pascal Context compared to baseline methods.

\modelname's preservation of high-resolution features within its superpixel representation thus proves to be a powerful asset for semantic segmentation, enabling detailed analyses of diverse image datasets.

\begin{table}[htb]
\caption{Semantic segmentation on ADE20K val split.}
\centering
\resizebox{\linewidth}{!}{
\begin{tabular}{lcccc}
\toprule
Method & \#Params & \#FLOPs & Pretrained & mIoU \\
\midrule
DeiT-S & 22M & 32G & \xmark & 20.1 \\
\modelname-S & 23M & 35G & \xmark & 23.1 \\
\midrule
DeiT-S & 22M & 32G & \cmark & 42.3 \\
\modelname-S & 23M & 35G & \cmark & 46.5 \\
\bottomrule
\end{tabular}
}
\vspace{-1em}
\label{tab:semantic_seg}
\end{table}

\begin{table}[htb]
\caption{Semantic segmentation on Pascal Conext val split.}
\centering
\resizebox{\linewidth}{!}{
\begin{tabular}{lcccc}
\toprule
Method & \#Params & \#FLOPs & Pretrained & mIoU \\
\midrule
DeiT-S & 22M & 27G & \xmark & 18.0 \\
\modelname-S & 23M & 30G & \xmark & 21.1 \\
\midrule
DeiT-S & 22M & 27G & \cmark & 48.3 \\
\modelname-S & 23M & 30G & \cmark & 51.2 \\
\bottomrule
\end{tabular}
}
\label{tab:semantic_seg_pascal}
\vspace{-1em}
\end{table}

\section{Conclusion}
\label{sec:conclusion}

In this work, we introduced SPFormer, a novel approach for feature representation in vision transformers, emphasizing superpixel representation. This method marks a shift from traditional pixel and patch-based approaches, offering three distinct advantages: explainability through semantic grouping of pixels, efficiency due to a reduced number of superpixels facilitating global self-attention, and robustness against image distortion. SPFormer not only demonstrates improved performance on ImageNet classification tasks but also excels in semantic segmentation, showcasing its versatility. Our results underscore the potential of superpixel-based methods in diverse vision tasks, paving the way for future exploration in this promising direction.

{
    \small
    \bibliographystyle{ieeenat_fullname}
    \bibliography{main}
}

\end{document}